\documentclass[letterpaper, 10 pt, journal, twoside]{IEEEtran}  

\IEEEoverridecommandlockouts                              

\usepackage{graphicx} 
\usepackage{amsmath} 
\usepackage{microtype}
\usepackage{hyperref}

\usepackage{fancyhdr}
\begin{document}

\title{Delta Descriptors: Change-Based Place Representation for Robust Visual Localization}

\author{Sourav Garg$^{1}$, Ben Harwood$^{2}$, Gaurangi Anand$^{3}$ and Michael Milford$^{1}$
\thanks{Manuscript received: Feb 24, 2020; Revised: May 13, 2020; Accepted: Jun 9, 2020.}
\thanks{This paper was recommended for publication by Sven Behnke upon evaluation of the Associate Editor and Reviewers' comments.
This work was supported by an AOARD grant: FA2386-19-1-4079.} 
\thanks{$^{1}$Sourav Garg and Michael Milford are with School of Electrical Engineering and Robotics, Australian Centre for Robotic Vision (ACRV) and the QUT Centre for Robotics (QCR), Queensland University of Technology, Australia.
        {\tt\footnotesize s.garg@qut.edu.au}
        }%
\thanks{$^{2}$Ben Harwood is with Department of Electrical and Computer Systems Engineering, Monash University, Australia.
        }%
\thanks{$^{3}$Gaurangi Anand is with School of Computer Science, Queensland University of Technology, Australia.}
\thanks{Digital Object Identifier (DOI): see top of this page.}
}


\maketitle
\thispagestyle{fancy}
\pagestyle{plain}

\begin{abstract}
Visual place recognition is challenging because there are so many factors that can cause the appearance of a place to change, from day-night cycles to seasonal change to atmospheric conditions. In recent years a large range of approaches have been developed to address this challenge including deep-learnt image descriptors, domain translation, and sequential filtering, all with shortcomings including generality and velocity-sensitivity. In this paper we propose a novel descriptor derived from tracking changes in \textit{any} learned global descriptor over time, dubbed \textit{Delta Descriptors}. Delta Descriptors mitigate the offsets induced in the original descriptor matching space in an unsupervised manner by considering temporal differences across places observed along a route. Like all other approaches, Delta Descriptors have a shortcoming - volatility on a frame to frame basis - which can be overcome by combining them with sequential filtering methods. Using two benchmark datasets, we first demonstrate the high performance of Delta Descriptors in isolation, before showing new state-of-the-art performance when combined with sequence-based matching. We also present results demonstrating the approach working with four different underlying descriptor types, and two other beneficial properties of Delta Descriptors in comparison to existing techniques: their increased inherent robustness to variations in camera motion and a reduced rate of performance degradation as dimensional reduction is applied. Source code is made available at \href{https://github.com/oravus/DeltaDescriptors}{https://github.com/oravus/DeltaDescriptors}.

\end{abstract}

\section{Introduction}

\IEEEPARstart{V}{isual} Place Recognition (VPR) is one of the key enablers for mobile robot localization and navigation. The earlier approaches to VPR predominantly relied on hand-crafted local (SIFT~\cite{lowe2004distinctive}, ORB~\cite{rublee2011orb}) and global (HoG~\cite{dalal2005histograms}, GIST~\cite{oliva2005gist}) image representation methods. The use of local features in BoW~\cite{sivic2003video} and VLAD~\cite{jegou2010aggregating} like encoding techniques based on visual vocabularies has been a popular choice for the task of global place retrieval (kidnapped robot problem). However, the lack of appearance robustness of underlying local features led to the development of appearance-invariant whole-image description techniques, combined with sequence-based matching~\cite{milford2012seqslam,naseer2014robust} in order to deal with extreme appearance variations.

Recent advances in deep learning have led to more robust counterparts to hand-crafted local and global feature representations like LIFT~\cite{yi2016lift}, DeLF~\cite{noh2017large}, NetVLAD~\cite{arandjelovic2016netvlad}, and LoST~\cite{garg2018lost}. Furthermore, GANs~\cite{goodfellow2014generative} based night-to-day image translation~\cite{anoosheh2018night}, feature fusion~\cite{hausler2019multi}, and teacher-student networks~\cite{sarlin2018bleveraging} have also been explored for VPR. Although achieving state-of-the-art performance, deep learning based image descriptors often suffer from the challenges of data bias that limits their utility for out-of-the-box operations. This is typically resolved by re-training or fine-tuning the CNN. However, this may not always be feasible for all the application scenarios like VPR: supervised learning would require multiple traverses of the new environment such that it captures variations in scene appearance and camera viewpoint. 

\begin{figure}
    \centering
    \includegraphics[scale=0.24]{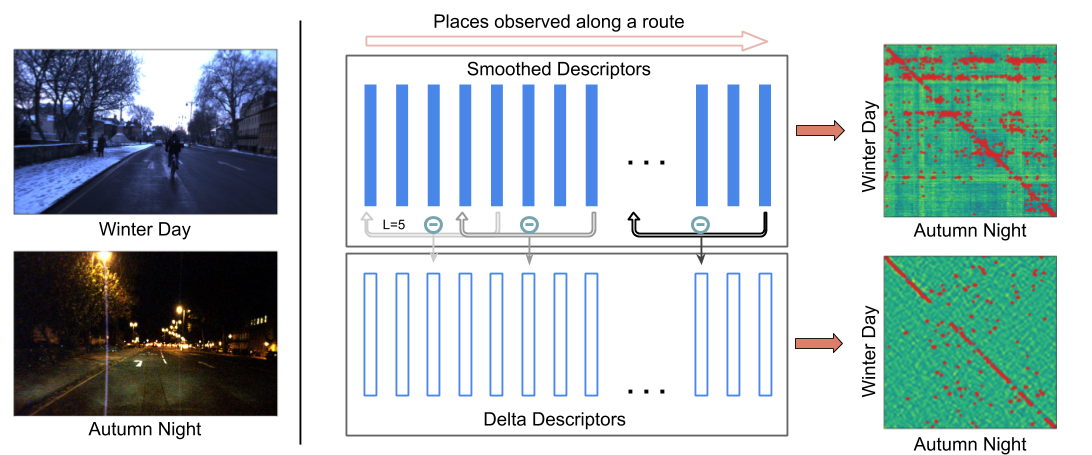}
    \caption{We propose Delta Descriptor, defined as a high-dimensional signed vector of change measured across the places observed along a route. Using a difference-based description, places can be effectively recognized despite significant appearance variations. The cosine distance matrix between the Winter Day and Autumn Night traverse of the Oxford Robotcar dataset~\cite{maddern20171} is displayed on the right with red markers indicating the predicted matches and the matrix diagonal representing its ground truth.}
    \label{fig:schematic}
\end{figure}

In this paper, we propose an unsupervised method for transforming existing deep-learnt image descriptors into change-based representations, dubbed \textit{Delta Descriptors}. Due to appearance variations in the environment during a revisit, the original descriptors tend to be offset by a margin, leading to an increased distance between descriptors belonging to the same place. Defined in a difference space~\cite{fukui2015difference}, Delta Descriptors implicitly deal with such offsets, resulting in a reduced distance between the same place descriptors. This is of particular importance to a mobile robot or autonomous vehicle operating in a new environment that may undergo significant variations in visual appearance during a repeated route traversal. Previously, difference-based approaches have been explored for recognizing objects~\cite{fukui2015difference}, faces~\cite{fukui2019discriminant} and actions~\cite{gatto2019tensor}. In this research, the concept of difference-based description is based on measuring the changes across an observed sequence of places which is repeatable across traverses. 

Using the proposed Delta Descriptors, we: 
\begin{itemize}
    \item establish that given a fixed sequential span, Delta Descriptors perform at par with sequence-based matching, and
    \item that they achieve state-of-the-art performance when used in conjunction with the sequence-based matching,
    \item show Delta Descriptors retain high performance when dimensional reduction techniques like PCA are used, in contrast to raw descriptors, especially in the presence of strong appearance variations,
    \item demonstrate their robustness to variations in camera motion both along and across the repeated traverses, unlike sequence-based matching methods which either require motion information or a sophisticated sequence-search approach, and
    \item provide insights into selecting sequential span sizes for calculating Delta Descriptors, the role of order in which places are observed, an investigative Multi-Delta Descriptor approach to deal with velocity variability, and image-level visualization of how the proposed descriptors aid in time-series pattern matching.
\end{itemize}

The paper is divided into the following sections: Section~\ref{sec:LitRev} discusses the prior literature and related work; Section~\ref{sec:Approach} describes the proposed approach for calculating Delta Descriptors; Section~\ref{sec:Exp} details the experimental setup including dataset description, system parameter estimation, and evaluation methods; Section~\ref{sec:Results} presents the results on the benchmark datasets and characterizes the proposed descriptor through a range of experiments; Section~\ref{sec:Disc} dives into the visualization of CNN activations of image regions for Delta Descriptors; and Section~\ref{sec:Conclusion} concludes the paper, highlighting potential ways to extend the current work in future. 

\section{Related Work}
\label{sec:LitRev}

\subsection{Hand-crafted and Deep-Learnt Descriptors}
The ability to represent images as a compact descriptor remains a key requirement for VPR. Broadly speaking, the purpose of these descriptors is to map images of a particular physical location into a lower dimensional representation. In the context of VPR, the goal of these mappings is to preserve a unique description of each location while removing changing information such as camera orientation, ambient lighting, mobile distractors and seasonal variations.

In the era of predominantly hand-crafted descriptors, the ORB~\cite{rublee2011orb} descriptor was designed to utilise a training set of local image patches. These patches were used to generate a set of binary tests that would maximise rotational invariance. While this learning procedure enabled the creation of more computationally efficient features, it also introduced the usual biases that come from training a model on a particular data set.

With the advent of deep learning, hand-crafted descriptors such as SIFT~\cite{lowe2004distinctive} have largely been replaced by learned descriptors such as LIFT~\cite{yi2016lift} and DeLF~\cite{noh2017large}. Learning these local patch-based representations end-to-end has again yielded improved descriptors, but has also increased the reliance on having training data that accurately models all aspects of the target domain. This data bias is also seen in learned global representations such as NetVLAD~\cite{arandjelovic2016netvlad} and AMOSNet~\cite{chen2017deep}.

\subsection{Sequence-based Representations}
A vast literature exists for spatio-temporal representation of video data with applications in action classification~\cite{girdhar2017actionvlad}, activity recognition~\cite{jalal2017robust}, person re-identification~\cite{wu20193}, dense video captioning~\cite{li2018jointly}, 3D semantic labelling~\cite{Song2016}, 3D shape completion~\cite{Dai2017a}, and dynamic scene recognition~\cite{feichtenhofer2014bags}. This has led to the emergence of 3D CNNs~\cite{karpathy2014large, tran2015learning} involving 3D convolutions to learn better spatio-temporal representations that are suited to the task at hand. However, most of the aforementioned tasks only require dealing with a limited number of classes unlike VPR where every other place is a unique observation. Furthermore, methods based on RNNs, LSTM networks~\cite{hochreiter1997long} and GRUs~\cite{chung2014empirical} tend to learn general patterns of how temporal information is ordered, for example, in the case of action or activity recognition. For VPR, such general patterns of order may not exist beyond the overlapping visual regions around a particular visual landmark. 

In the context of VPR, there have been some attempts towards developing robust spatio-temporal or sequence-based representations. In \cite{johns2011place}, authors learnt spatio-temporal landmarks based on SURF features, but these local features needed to be independently tracked on frame-to-frame basis. In \cite{nguyen2013spatio}, authors proposed a bio-inspired place recognition method that used environment-specific discriminative training of different Long-Term Memory (LTM) cells. \cite{facil2019condition} explored three novel techniques: Descriptor Grouping, Descriptor Fusion, and Recurrent Descriptors, to accrue deep features from multiple views of the scene across the route. \cite{garg2019look} proposed a topometric spatio-temporal representation of places using monocular depth estimation, mainly focused on recognizing places from opposing viewpoints. \cite{volkov2015coresets} proposed coresets-based visual summarization method for efficient hierarchical place recognition. Doing away with a compact representation, \cite{zhang2016robust} used a variety of image descriptors to represent groups of images and formulated sequence-searching as an optimization problem.

\subsection{Sequence-based Matching}
Although sequence-based representations are not that common in VPR literature, use of sequence-based matching has been extensively explored for VPR. Such methods leverage sequential information \textit{after} computing the place matching scores where place representations are typically based on a single image. This leads to enhanced VPR performance~\cite{cummins2008fab}, particularly in cases where perceptual aliasing is very high, for example, dealing with extreme appearance variations caused by day-night and seasonal cycles, using methods like SeqSLAM~\cite{milford2012seqslam} and SMART~\cite{pepperell2014all}. The follow up work in this direction is comprised of a number of methods that deal with camera velocity sensitivity~\cite{naseer2014robust, vysotska2016lazy} or velocity estimation~\cite{garg2017straightening}. More recent work includes using temporal information and diffusion process within graphs~\cite{zhang2019graph}, multi-sequence maps based VPR~\cite{vysotska2019effective}, and trajectory attention-based learning for SLAM~\cite{parisotto2018global}. In this paper, we use a simplified sequence-based matching technique mainly to analyse performance dynamics in using temporal information in two very different ways: sequential representation and sequential matching.

\subsection{Difference-based Representations}
The concept of using difference-based representation has been explored in a few different ways. \cite{fukui2015difference} proposed Generalized Difference Subspace (GDS) as an extension of a difference vector for analyzing shape differences between objects. \cite{fukui2019discriminant} proposed a novel discriminant analysis based on GDS demonstrating its utility as discriminative feature extractor for face recognition. Recently, \cite{gatto2019tensor} extended the concept of GDS to tensors for representing and classifying gestures and actions. \cite{zhu2016sparse} used difference subspace analysis to maximize inter-class discrimination for effective face recognition as an alternative approach to improving representation ability of samples. \cite{tseng2012human} proposed a human action recognition method based on difference information between spatial subspace of neighboring frames. \cite{jalal2017robust} used sum of depth difference between consecutive frames to discriminate moving/non-moving objects in order to detect humans. Our proposed method is based on descriptor difference and in essence solves the problem of recognition in a similar way as the GDS-based methods. In particular, Delta descriptors enable dealing with the offset that occurs in the deep-learnt place representations when a robot is operating in new environments under significantly different environmental conditions.

\begin{figure}
    \centering
    \includegraphics[scale=0.4]{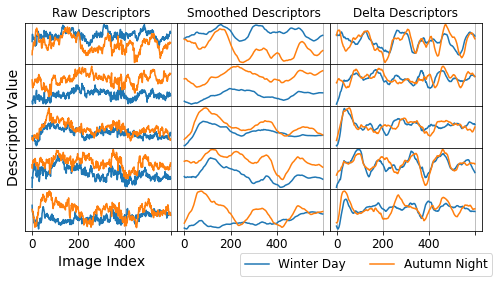} 
    \caption{For five selected descriptor dimensions (across rows), time-series of observed places for two traverses from Oxford Robotcar dataset are displayed for $\ell_{2}$-normalized Raw (left), Smoothed (middle) and Delta Descriptors (right). The latter two are computed using a $64$-frame window. The time-series pairs (blue and orange) should ideally be well aligned with each other.}
    \label{fig:visNVLAD}
\end{figure}

\section{Proposed Approach}
\label{sec:Approach}

A vast majority of existing place representation methods use single-image based descriptors for place recognition, typically followed by sequential matching or temporal filtering performed over place-matching scores. In this paper, we follow an alternative approach to place representation and propose Delta Descriptors that leverage the sequential information by measuring the \textit{change} in descriptor as different places are observed over time. We hypothesize that these changes are both unique and consistent across multiple traverses of the environment, despite significant variations in scene appearance. In particular, measuring the change inherently ignores the data bias of the underlying deep-learnt image descriptors, elevating the latter's utility under diverse environment types and appearance conditions.

In this section, we first highlight key observations from the time-series of existing state-of-the-art image descriptors, then define and formulate the Delta Descriptors, and finally, describe an alternative convolutions-based approach to compute the Delta Descriptors more efficiently.

\paragraph{Key Observations} In the context of VPR for a mobile robot, images are typically captured as a data stream and converted into high-dimensional descriptors. We consider this stream of image descriptors as a multi-variate time-series. For some of the descriptor dimensions\footnote{The dimension indices were selected using the method described in Section~\ref{sec:VisAct} using NetVLAD descriptors.}, Figure~\ref{fig:visNVLAD} shows pairs of time-series for first $600$ images (with $0.5$ meters frame separation) from two different traverses of Oxford Robotcar dataset captured under day and night time conditions respectively. For the Raw descriptors, it can be observed that the consecutive values in the time-series tend to vary significantly even though the adjacent frames have high visual overlap. Moreover, the local variations in the descriptor values are not consistent across the traverses albeit the global time-series patterns appear repeatable. As the underlying deep-learnt image descriptors (in most cases) are not trained to be stable against slight perturbations in camera motion or for ignoring the dynamic objects in the scene, such local variations are an expected phenomenon.

\paragraph{Defining Delta Descriptors} With these observations, we define delta descriptor, $\mathbf{\Delta}_t$, as a high-dimensional signed vector of change measured across a window of length $l$ over a \textit{smoothed} multi-variate time-series, $\mathbf{X}_t \in \mathcal{R}^D$, where $t$ represents the time instant for an observation of a place along a route in the form of an image descriptor. To be more specific, we have

\begin{equation}
    \mathbf{\Delta}_t = \overline{\mathbf{X}}_{t+l/2} - \overline{\mathbf{X}}_{t-l/2}
    \label{eq:DD}
\end{equation}

\noindent where $\overline{\mathbf{X}}_t$ represents the smoothed signal obtained from a rolling average of the time series $\mathbf{X}_t$:

\begin{equation}
    \overline{\mathbf{X}}_t = \sum_{t'=t-l/2}^{t+l/2} \mathbf{X}_{t'}/l
    \label{eq:smooth}
\end{equation}

The middle and the right graphs in Figure~\ref{fig:visNVLAD} show the smoothed time-series and the corresponding Delta Descriptor respectively. It can be observed that the proposed descriptors are much more aligned than the baseline ones, getting rid of the offset in their original values.

\paragraph{Simplified Implementation} The formulation for Delta Descriptors presented above is suitable for understanding and visualizing the time-series patterns. However, Equation~\ref{eq:DD} and \ref{eq:smooth} can be simplified to a convolution-based calculation of the proposed descriptors:

\begin{equation}
    \mathbf{\Delta} = \mathbf{X} \ast \mathbf{W}
    \label{eq:conv}
\end{equation}

\noindent where convolutions are preformed along the time axis of the baseline descriptor, independently per dimension using a 1D convolutional filter $\mathbf{W} = (w_1, w_2, \dots, w_L)$ defined as a vector of length $L=2l$:

\begin{equation}
    w_i =
    \begin{cases}
        -1/l, & \text{if } i \leq L/2,\\
        \phantom{-}1/l, & \text{otherwise}.
    \end{cases}
\end{equation}

For performing visual place recognition, the proposed descriptors are matched using cosine distance\footnote{If Euclidean distance is used for this purpose, Delta Descriptors would need to be $\ell_{2}$-normalized.}. However, for visualization purposes as in Figure~\ref{fig:visNVLAD}, individual descriptors are $\ell_{2}$-normalized, knowing that the Euclidean distance between pairs of normalized descriptors is proportional to the cosine distance between their un-normalized counterparts.

\section{Experimental Setup}
\label{sec:Exp}

\subsection{Datasets}
We used subsets of two different benchmark datasets to conduct experiments: Oxford Robotcar~\cite{maddern20171} and Nordland~\cite{sunderhauf2013we}. Repeated route traverses from these datasets exhibit significant variations in scene appearance due to changes in environmental conditions caused by time of day and seasonal cycles.

\paragraph{Oxford Robotcar} This dataset is comprised of $10$ km traverses of urban regions of Oxford city captured under a variety of environmental conditions. We used the forward-facing camera imagery from the first 1 km of three different traverses, referred to as Summer Day, Winter Day and Autumn Night in this paper\footnote{corresponding to 2014-07-14-14-49-50, 2014-11-14-16-34-33, 2015-02-03-08-45-10 respectively.}. For all three traverses, we used a constant frame separation of $0.5$ meters, leading to a database of around $2000$ images per traverse.

\paragraph{Nordland} This dataset comprises $728$ km train journey across vegetative open environment from Nordland captured under four seasons. We used the first $1750$ images (out of $35768$) from the Summer and Winter traverse after skipping the first $250$ frames where the train was stationary.

\subsection{Parameter: Sequence Length}
The concept of Delta descriptors is based on measuring the changes in visual information in the form of places observed during a traverse which are then expected to be preserved across subsequent traverses. When using a very short sequence length, such changes can be erratic due to high visual overlap between adjacent frames. This is partly due to the unstable response output from the underlying CNN as it is not trained to produce smooth variation in descriptor for small variations in camera motion, as shown in Figure~\ref{fig:visNVLAD}. In order to choose the sequence length parameter for our experiments, we used the relative distribution of cosine distance between descriptors, obtained by matching a dataset with itself.

\begin{figure}
    \centering
    \includegraphics[scale=0.32]{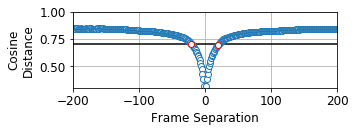}
    \includegraphics[scale=0.32]{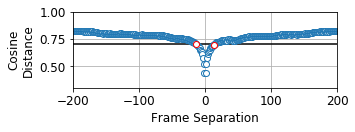}
    \caption{Median of cosine distance between descriptors of neighboring image frames is plotted against their frame separation for Oxford Winter Day (left) and Nordland Summer (right) traverses to estimate the sequence length parameter for calculating Delta Descriptors.}
    \label{fig:paramSeqL}
\end{figure}

Figure~\ref{fig:paramSeqL} shows these distributions for Winter Day traverse from the Oxford Robotcar dataset and Summer traverse from the Nordland dataset, where cosine distance is plotted against frame separation as the median value computed across the whole traverse. We found that using a fixed cosine distance threshold of $0.7$ (black horizontal line), a minimum bound on the sequence length parameter can be directly estimated from these distributions (shown with red circles), making sure that the place observation has changed sufficiently enough to robustly measure the changes in descriptor values. Using this method, the sequential span was found to be $38$, $36$ and $57$ for the Oxford datasets: Winter Day, Summer Day and Autumn Night. For the Nordland Summer and Winter traverses, these values were estimated to be $14$ and $8$. Hence, as a lower bound, we compute Delta Descriptors using a fixed sequence length of $64$ and $16$ frames for all the traverses of the Oxford Robotcar and the Nordland dataset respectively.

\subsection{Evaluation}
We used Precision-Recall (PR) curves to measure VPR performance. For a given localization radius, precision is defined as the ratio of correct matches to total matches retrieved and recall is defined as the ratio of correct matches to total possible true matches. For datasets used in this paper, a true match exists for every query image. A match for a query is retrieved only when its cosine distance is less than a threshold which is varied to generate the PR curves. We present PR curves for two different values of localization radii: $10$ and $40$ \textit{meters} for the Oxford Robotcar dataset and $2$ and $10$ \textit{frames} for the Nordland dataset. For some of the experiments, we also report Precision at $100\%$ Recall which is useful for re-ranking based hierarchical localization pipelines~\cite{garg19Semantic,mur2017orb}.

\subsection{Comparative Study}
We used the state-of-the-art single image-based descriptor NetVLAD~\cite{arandjelovic2016netvlad} as a baseline in the results, represented as \textit{Raw Descriptors}. \textit{Delta Descriptors} were calculated using Equation~\ref{eq:conv} with NetVLAD as the underlying descriptor (see Section~\ref{sec:MoreDescs} for experiments using different underlying descriptor). As the proposed Delta descriptors use sequential information, we also compare them against a naive sequential representation of NetVLAD, achieved by smoothing the baseline descriptors using Equation~\ref{eq:smooth}, represented as \textit{Smoothed Descriptors}. Furthermore, we also consider the orthogonal approach to utilizing sequences for VPR that is based on sequential aggregation of match scores, typically obtained by comparing single image descriptors. For this, we use a simplified version of sequence matching which is similar to~\cite{milford2012seqslam} but only aggregates match scores along a straight line without any velocity searching~\cite{garg2020fast}. We refer to this as \textit{SeqMatch} in the results and use it on top of the Raw, Smoothed and Delta descriptors.

\section{Results}
\label{sec:Results}
In this section, we first present the benchmark comparisons on three pairs of route traverses using two different datasets. Then, we demonstrate the performance effects of PCA transformation, data shuffling within a traverse, variations in camera motion and sequential-span searching using Multi-Delta Descriptors.

\newcommand{\scaleOne}{0.3}
\begin{figure}
    \centering
    \begin{tabular}{cc}
    \multicolumn{2}{c}{\includegraphics[scale=\scaleOne,trim={0 3.75cm 0 3.65cm},clip]{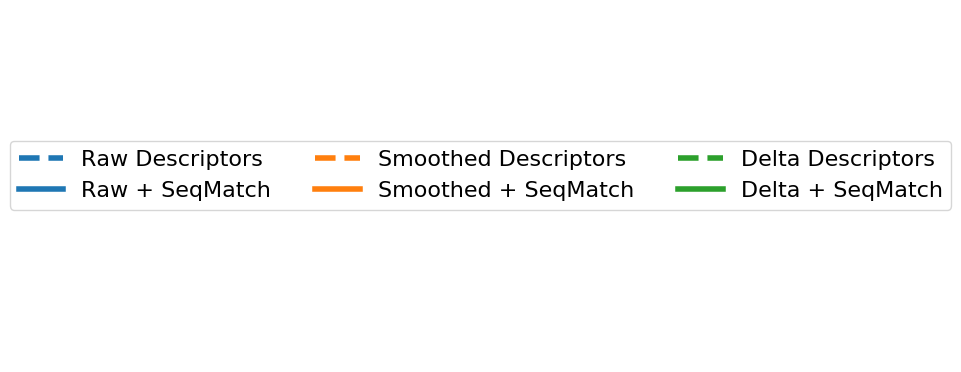}} \\
    \includegraphics[scale=\scaleOne]{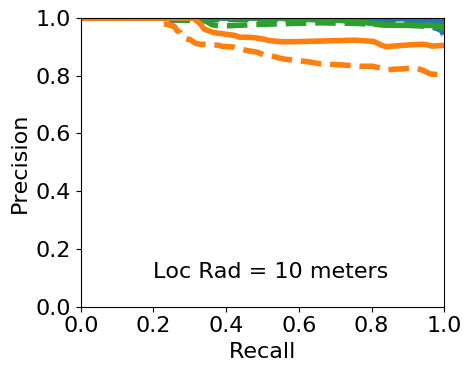} & 
    \includegraphics[scale=\scaleOne]{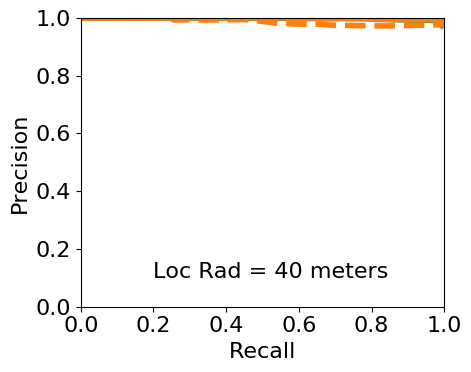} \\ 
    \multicolumn{2}{c}{(a) \textit{Oxford} Winter Day vs Summer Day} \\  
    \includegraphics[scale=\scaleOne]{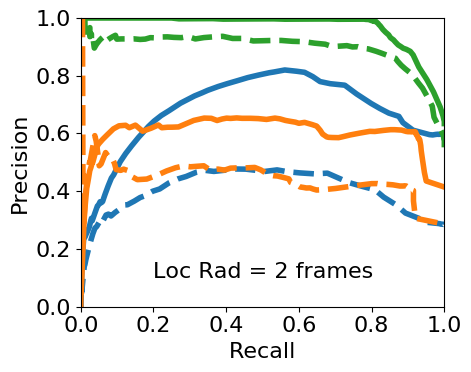} & 
    \includegraphics[scale=\scaleOne]{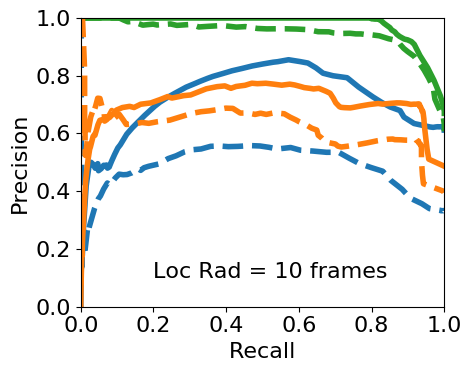} \\ 
    \multicolumn{2}{c}{(b) \textit{Nordland} Summer vs Winter} \\     
    \includegraphics[scale=\scaleOne]{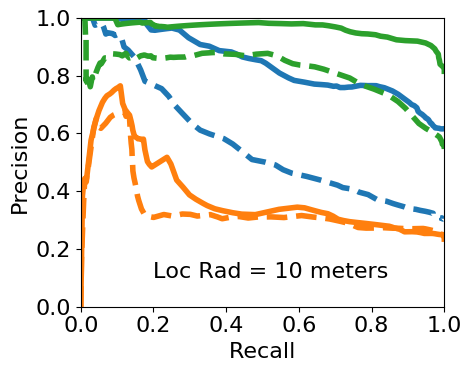} & 
    \includegraphics[scale=\scaleOne]{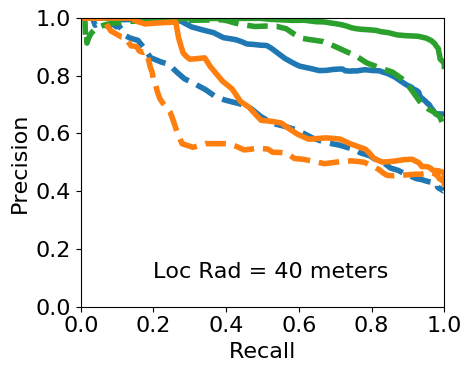} \\ 
    \multicolumn{2}{c}{(c) \textit{Oxford} Autumn Night vs Winter Day}  
    \end{tabular}
    \caption{Precision-Recall performance comparison on three pairs of traverses from the Oxford Robotcar and Nordland datasets.}
    \label{fig:mainRes}
\end{figure}

\subsection{Sequential Representations and/vs Sequential Matching}
Figure~\ref{fig:mainRes} shows the results for two pairs of traverses from the Oxford Robotcar dataset and one pair from the Nordland dataset. The sequential span L was set to $64$ frames ($32$ meters) for the former and $16$ frames for the latter. 

As a general trend, it can be observed that Delta Descriptors outperform both Raw and Smoothed descriptors, leading to a much higher recall in the high-precision region. While performance levels are observed to be saturated for the day-day comparison across different seasons in an urban city environment (Figure~\ref{fig:mainRes}a), the absolute performance of raw descriptors is observed to be quite low when such appearance variations occur in natural open environment (Figure~\ref{fig:mainRes}b). Such low performance might be due to lack of generalization ability of the NetVLAD descriptors. Thus, in contrast to requiring supervised fine-tuning or re-training, Delta Descriptors tend to mitigate the issue in a completely unsupervised manner.

In Figure~\ref{fig:mainRes}b, it can also be observed that even using sequence matching on top of raw descriptors (solid blue) cannot achieve performance similar to that attained using Delta Descriptors without sequence matching (dashed green). This particularly highlights the effectiveness of sequence-based place representation as opposed to sequence-based aggregation/filtering of matching scores, given a fixed sequential span.

Figure~\ref{fig:mainRes}c shows performance trends for a more difficult scenario combining the challenges of both seasonal (autumn/winter) and time-of-day (day/night) variations. It can be observed that even without using sequence matching on top, Delta Descriptors perform on par with the Raw+SeqMatch combination, except when considering recall at $100\%$ precision. We observed that the averaging operation in both Smoothed and Delta Descriptors leads to some loss of precision, particularly apparent when the smoothing window size is larger than the considered localization radius. This loss in precision can typically be mitigated by sequential matching where the spurious matching scores are averaged out. In Figure~\ref{fig:mainRes}, the benefits of sequential matching can be consistently observed in all the results where it not only improves the overall recall performance but simultaneously maintains a high precision level. It is worth noting that a subsequent geometric verification step, commonly employed in hierarchical localization pipelines~\cite{cummins2008fab, garg19Semantic, mur2017orb}, can further improve the precision performance.

With the use of sequentially-ordered information, Delta descriptors are able to neglect the offsets in the baseline descriptors that occur due to significant variations in scene appearance. This leads to superior performance as compared to the raw descriptors, especially under the challenging scenario of day vs night (see Figure~\ref{fig:mainRes}c). Furthermore, it can be observed that descriptor smoothing applied naively to the baseline descriptors is of a limited use. This is due to the dilution of discriminative information within the descriptors as they all get closer to the mean of the data. Finally, it can be observed that sequence matching enhances the performance of Delta descriptors more than the raw descriptors, indicating the better representation ability of the former.

\subsection{Dimension Reduction via PCA}
\begin{figure}
    \centering
\begin{tabular}{cc}
    \includegraphics[scale=\scaleOne]{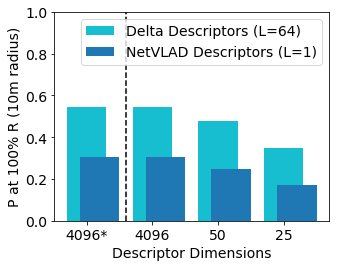} &
    \includegraphics[scale=\scaleOne]{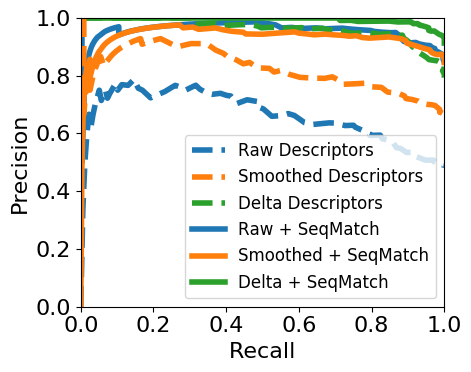} \\
    (a) & (b)
\end{tabular}
    \caption{(a) Effect of PCA transformation on performance of Raw NetVLAD and Delta Descriptors using Oxford Robotcar Day-Night traverses. * means no PCA transformation. (b) Performance comparisons using Nordland Summer and Winter traverses with random shuffling of data such that the order of places is preserved across the traverses but not within.}
    \label{fig:res_PCA}
\end{figure}

Image descriptors obtained through CNNs are typically high-dimensional. For global retrieval tasks, computational complexity is often directly related to the descriptor dimension size. Therefore, dimension reduction techniques like PCA are commonly employed~\cite{arandjelovic2016netvlad,garg2018don't,lowry2016supervised,schubert2020unsupervised}. However, this can lead to significant performance degradation due to extreme variations in scene appearance as the variance distribution in the original descriptor space may not be repeatable. In Figure~\ref{fig:res_PCA}a, we show the effect of PCA-based dimension reduction on the performance of Raw NetVLAD and Delta descriptors using the Oxford Robotcar day-night traverses. It can be observed that the proposed Delta Descriptors are robust to dimension reduction techniques like PCA: even retaining only $50$ principal components does not degrade the performance much. On the other hand, the baseline NetVLAD descriptors suffer significant performance drop with PCA even when all $4096$ components are retained, highlighting its sensitivity to data centering.

\subsection{Order of Place Observations}
The concept of Delta descriptors is based on the sequential order of changes in visual information. In the context of VPR, sequentially observed places typically have some visual overlap which affects the overall dynamics of performance when considering either sequential representation or sequential matching. We consider another scenario where both the reference and the query data are shuffled such that the order of images is preserved across traverses but there is no visual overlap between adjacent frames. For this, we used the Nordland dataset and sampled every $100^{th}$ image out of $35768$ images and then performed the shuffling. In Figure~\ref{fig:res_PCA}b, we can observe that even with a sequential span of $2$ frames (lacking visual overlap) and localization radius of $1$ frame, additional information in the form of sequences can be better utilized with sequence-based descriptors than sequence-based match-score aggregation of single image descriptors, while their combination achieves even higher performance. Furthermore, this experiment also indicates that the concept of Delta Descriptors is applicable in general to describing and matching ordered observations, irrespective of whether or not the adjacent elements are related to each other.

\subsection{Camera Motion Variations \& Multi-Delta Descriptors}
In our previous experiments, we used a constant frame spacing between the reference and the query traverses ($0.5$ meters for Oxford Robotcar). In practice, camera velocity may change both within and across the repeated traverses of the environment. In order to observe the effect of such variations on the VPR performance, we conducted another experiment using the first $8600$ images\footnote{only every $5^{th}$ frame was considered, leading to the dataset size of $1720$ images each. Note that this does not affect the camera velocity and was only done to reduce the processing time.} ($\sim$1 km) from the Winter Day and Autumn Night traverses without any data pre-processing, that is, without motion-based keyframe selection.

For this study, we used a sequential span of $64$ frames both for computing Delta Descriptors and sequence matching. Furthermore, in order to deal with variable motion both across and within the traverses, we present a preliminary investigation into sequential-span searching using a Multi-Delta Descriptor approach. To achieve this, for both the reference and the query data, multiple Delta descriptors are computed using a range of sequential spans: $\{30,40,50,60\}$ frames in this case. The match value for any given image pair is calculated as the minimum of the cosine distance over all the possible combinations of sequential spans ($16$ here) used for computing multiple Delta Descriptors.

\begin{figure}
    \centering
\begin{tabular}{cc}
    \multicolumn{2}{c}{\includegraphics[scale=\scaleOne,trim={0 3.75cm 0 3.65cm},clip]{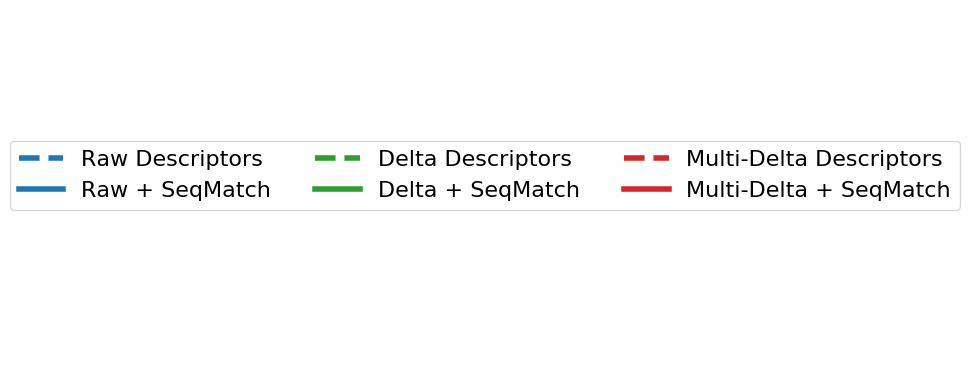}} \\
    \includegraphics[scale=\scaleOne]{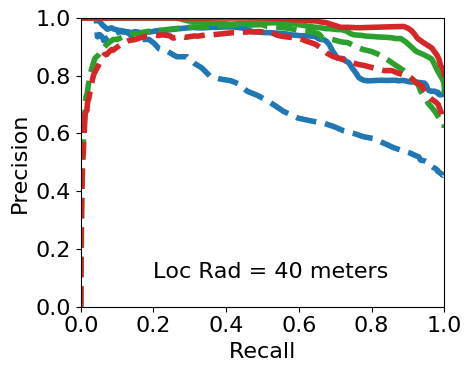} &
    \includegraphics[scale=\scaleOne]{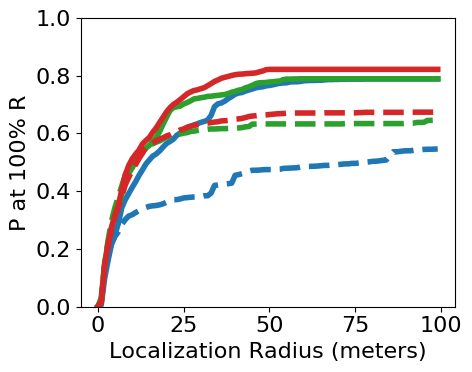}
\end{tabular}
    \caption{Performance comparison using Oxford Autumn Night and Winter Day traverses without any pre-processing of camera motion.}
    \label{fig:res_noGPS}
\end{figure}

In Figure~\ref{fig:res_noGPS}, we can observe that even without any data pre-processing (keyframe selection), state-of-the-art performance is achieved by the Delta Descriptors + SeqMatch combination. As we do not use any local velocity-search technique (originally proposed in \cite{milford2012seqslam}) in our sequential matching method, performance contribution of SeqMatch is less due to velocity variations as compared against the constant-velocity experiments in Figure~\ref{fig:mainRes}c. However, the effect of local variations in camera velocity is less detrimental for Delta descriptors, leading to superior absolute performance even without any sequential matching. Finally, it can be observed that the Multi-Delta Descriptor approach further improves the state-of-the-art performance. This also emphasizes the highly discriminative nature of difference-based description that enables accurate match searching within a given range of sequential spans, potentially leading to its applications beyond \textit{exact} repeats of route traversals.

\section{Discussion}
\label{sec:Disc}

\subsection{Visualizing Variations in Activations}
\label{sec:VisAct}
In order to visualize how Delta descriptors utilize the difference-based sequential information, we used a Global Max Pooling (GMP) based image descriptor through which image region activations can be directly interpreted which is not trivial for VLAD pooling of NetVLAD. GMP descriptors are extracted from the final \textit{conv} layer of ResNet-50~\cite{he2016deep} and Delta Descriptors are calculated using them for Oxford's Winter Day and Summer Day traverses for this experiment.

For visualization purpose, using the ground truth for place matches, dimensions of the GMP descriptor were ranked in order to observe only those which contributed the most to the performance. This was achieved by taking an element-wise product of a known matching pair of descriptors and sorting them in the order of the product value (a higher value indicates that both the descriptors had a similar high activation). This process was repeated for all the pairs of descriptors ($\sim$2000) and the dimensions that repeatedly ranked higher (higher product value) were selected for visualization. 

In Figure~\ref{fig:visGMP} (a) and (c), GMP descriptor dimension index $1501$ is used. The graph in Figure~\ref{fig:visGMP}a shows the variation of descriptor values along the route for Raw, Smoothed, and Delta descriptors\footnote{All the descriptors are $\ell_{2}$-normalized independently to aid visualization in Euclidean space.} (from top to bottom). For image indices in the range of $400-800$, both the raw and the smoothed values do not align well across the traverses but are relatively closer in the Delta Descriptor space. Figure~\ref{fig:visGMP}c displays the $550^{th}$ image from the Winter (left) and the Summer day traverse (right) where mask color indicates the activation values that increase from blue to green and then to red. It can be observed that the activations for the image from the Summer traverse are lower than that for the Winter traverse due to different lighting conditions around the visual landmark, leading to an increased distance in the original descriptor space. However, as the Delta descriptor only considers changes \textit{within} a traverse, its descriptor values still remain consistent throughout, even though the absolute activation values are lower in one of the traverses. 

\subsection{Delta based on Different Underlying Descriptors}
\label{sec:MoreDescs}
For the Nordland (Winter vs Summer) dataset, Figure~\ref{fig:visGMP}b shows performance comparison between the Raw and Delta descriptors computed using four different underlying descriptors: NetVLAD, GMP (ResNet50), AMOSNet (fc7)~\cite{chen2017deep} and HybridNet (fc7)~\cite{chen2017deep}. It can be observed that irrespective of the underlying descriptor choice, Delta Descriptors lead to consistent performance gains with significant improvements for viewpoint-based descriptors (AMOSNet and HybridNet).

\newcommand{\scaleVisGMP}{0.08}
\begin{figure}
    \centering
    \begin{tabular}{cc}
    \includegraphics[scale=0.25]{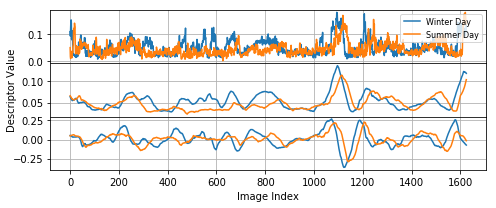} &
    \includegraphics[scale=\scaleOne]{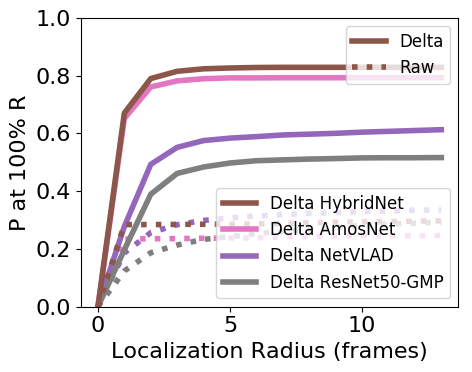} \\
    (a) & (b) \\
    \multicolumn{2}{c}{\includegraphics[scale=\scaleVisGMP]{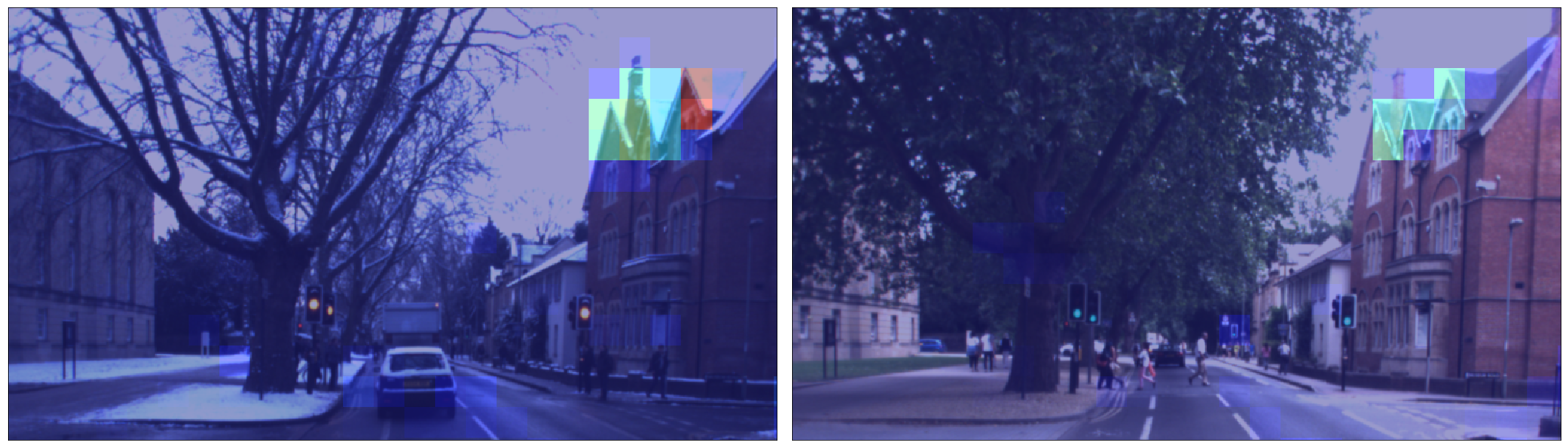}} \\
    \multicolumn{2}{c}{(c)} \\
    \end{tabular}
    \caption{(a) Time-series of GMP descriptor of images observed along a route for two different traverses. (b) Performance comparison using different underlying descriptors (Raw) to compute Delta Descriptor. (c) 550th image index from the Winter Day and Autumn Day traverse with colored activation masks overlaid (green and red correspond to 20 and 50 respectively).}
    \label{fig:visGMP}
\end{figure}

\section{Conclusion and Future Work}
Visual place recognition under large appearance changes is a difficult task. Existing deep-learnt global image description methods do enable effective global image retrieval. However, when operating in new environments where appearance conditions vary drastically due to day-night and seasonal cycles, these methods tend to suffer from an offset in their image description. Our proposed Delta Descriptors are defined in a difference space, which is effective at eliminating description offsets seen in the original space in a completely unsupervised manner. This leads to a significant performance gain, especially for the challenging scenario of day-night VPR. When considering a given sequential span, we have demonstrated that Delta Descriptors achieve state-of-the-art results when combined with sequential matching. This performance is a strong indicator of the robust representation ability given by Delta Descriptors. Finally, we have presented results for a range of experiments that show the robustness of our method when handling PCA-based dimensional reduction and variations in camera motion both along and across the repeated route traversals.

Our current work can be extended in several ways including estimating the descriptor transformation (offsets) on the fly, learning what visual landmarks are more suited to track changes, and measuring changes independently but simultaneously for different descriptor dimensions. In particular, it would be interesting to see how a framework that learns underlying descriptors would change its behaviour if optimized for place recognition performance using the subsequent Delta Descriptors. The concept of using a difference space~\cite{fukui2015difference} is not well-explored in the place recognition literature but is a promising avenue for future research applied to other similar problems where inferring or learning the changes might be more relevant than the representation itself~\cite{fukui2019discriminant, gatto2019tensor, zhu2016sparse}. We believe that our research contributes to the continued understanding of deep-learnt image description techniques and opens up new opportunities for developing and learning robust representations of places that leverage spatio-temporal information.

\label{sec:Conclusion}






\bibliographystyle{IEEEtran}
\bibliography{reflist,moreRefs}

\end{document}